\documentclass[lettersize,journal]{IEEEtran}
\usepackage{amsmath,amsfonts, amssymb}
\usepackage{algorithmic}
\usepackage{array}
\usepackage[caption=false,font=normalsize,labelfont=sf,textfont=sf]{subfig}
\usepackage{textcomp}
\usepackage{stfloats}
\usepackage{url}
\usepackage{verbatim}
\usepackage{graphicx}
\usepackage{cite}
\usepackage{algorithm}
\usepackage{xcolor}
\usepackage{arydshln}
\usepackage{booktabs}
\usepackage{tabularx}
\usepackage{caption}
\usepackage{nicefrac}
\usepackage{subfig}
\usepackage[utf8]{inputenc} 
\usepackage[T1]{fontenc}    
\usepackage{microtype}      
\usepackage{paralist}
\usepackage{pdfpages}
\usepackage{multirow}

\def\BibTeX{{\rm B\kern-.05em{\sc i\kern-.025em b}\kern-.08em
    T\kern-.1667em\lower.7ex\hbox{E}\kern-.125emX}}
\usepackage{balance}

\IEEEoverridecommandlockouts

\renewcommand{\IEEEbibitemsep}{0pt plus 0.5pt}
\makeatletter
\IEEEtriggercmd{\reset@font\normalfont\fontsize{6.9pt}{7.40pt}\selectfont}
\makeatother
\IEEEtriggeratref{1}
\begin{document}
\title{ESGNN: Towards Equivariant Scene Graph Neural Network for 3D Scene Understanding}
\author{\IEEEauthorblockN{Quang P.M. Pham\IEEEauthorrefmark{2}, Khoi T.N. Nguyen\IEEEauthorrefmark{2},\\
    Lan C. Ngo, Truong Do, Truong Son Hy \IEEEauthorrefmark{1}\\}
\IEEEauthorblockA{\textit{College of Engineering and Computer Science, VinUniversity, Vietnam \\
Department of Mathematics and Computer Science, Indiana State University, US \\
Email: \{20quang.ppm\IEEEauthorrefmark{2},  20khoi.ntn\IEEEauthorrefmark{2}, 20lan.nc, truong.dt\}@vinuni.edu.vn, 
TruongSon.Hy@indstate.edu}} 

\thanks{\IEEEauthorrefmark{2}: These authors contributed equally in this work}
\thanks{\IEEEauthorrefmark{1}: Corresponding author}
}


\maketitle

\begin{abstract}
Scene graphs have been proven to be useful for various scene understanding tasks due to their compact and explicit nature. However, existing approaches often neglect the importance of maintaining the symmetry-preserving property when generating scene graphs from 3D point clouds. This oversight can diminish the accuracy and robustness of the resulting scene graphs, especially when handling noisy, multi-view 3D data. This work, to the best of our knowledge, is the first to implement an Equivariant Graph Neural Network in semantic scene graph generation from 3D point clouds for scene understanding. Our proposed method, ESGNN, outperforms existing state-of-the-art approaches, demonstrating a significant improvement in scene estimation with faster convergence. ESGNN demands low computational resources and is easy to implement from available frameworks, paving the way for real-time applications such as robotics and computer vision.
\end{abstract}


\section{Introduction} \label{sec:intro}
Holistic scene understanding serves as a cornerstone for various applications across fields such as robotics and computer vision \cite{kim2024adaptive, LiCVPR, 10484453}. Scene graphs, which utilize Graph Neural Network (GNN), offer a lighter alternative to 3D reconstruction while still being capable of capturing semantic information about the scene. As such, scene graphs have recently gained more attention in the robotics and computer vision fields \cite{9661322}. For scene graph representation, objects are treated as nodes, and the relationships among them are treated as edges.

Recent advancements in scene graph generation have transitioned from solely utilizing 2D image sequences to incorporating 3D features such as depth camera data and point clouds, with the latest approaches, like \cite{wu2023incremental, Wald2020, wu2021scenegraphfusion}, leveraging both 2D and 3D information for improved representation. However, these methods overlook the symmetry-preserving property of GNNs, which potentially cause scene graphs' inconsistency, being sensitive to noisy and multi-view data such as 3D point clouds. Hence, this work adopts E(n) Equivariant Graph Neural Network \cite{pmlr-v139-satorras21a}'s Convolution Layers with Feature-wise Attention mechanism \cite{wu2021scenegraphfusion} to create \textbf{E}quivariant \textbf{S}cene \textbf{G}raph \textbf{N}eural \textbf{N}etwork (ESGNN). This approach ensures that the resulting scene graph remains unaffected by rotations and translations, thereby enhancing its representation quality. Additionally, ESGNN requires fewer layers and computing resources compared to Scene Graph Fusion (SGFN) \cite{wu2021scenegraphfusion}, while achieving higher accuracy scores with fewer training steps.

In summary, our contributions include:
\begin{compactitem}
    \item We, to the best of our knowledge, are the first to implement Equivariant GNN in generating semantic scene graphs from 3D point clouds for scene understanding.
    \item Our method, named ESGNN, outperforms state-of-the-art methods, achieving better accuracy scores with fewer training steps.
    \item We demonstrate that ESGNN is adaptive to different scene graph generation methods. Furthermore, there is significant potential to explore the integration of equivariant GNNs for scene graph representation, with considerable room for future improvement.
\end{compactitem}

\section{Overall Framework} \label{sec:background}

\begin{figure*}[htbp]
    \centering
    \includegraphics[width=\textwidth]{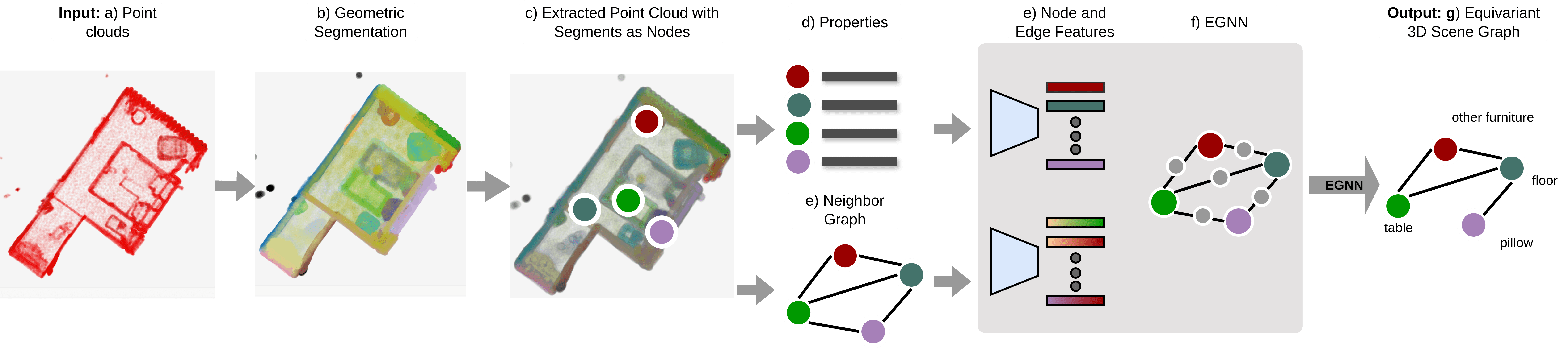}
    \caption{Overview of the proposed Equivariant Scene Graph framework. Our approach takes a sequence of point clouds a) as input to generate a geometric segmentation b). Subsequently, the properties of each segment and a neighbor graph between segments are constructed. The properties d) and neighbor graph e) of the segments that have been updated in the current frame c) are used as the inputs to compute node
and edge features f) and to predict a 3D scene graph g). }
    \label{fig:framework}
\end{figure*}

\textit{Problem Formulation:} The Semantic Scene Graph is denoted as $\mathcal{G}_s=(\mathcal{V}, \mathcal{E})$, where $\mathcal{V}$ and $\mathcal{E}$ represent sets of entity nodes and directed edges, respectively. In this case, each node $v_i \in \mathcal{V}$ contains an entity label $l_i \in L$, a point cloud $\mathcal{P}_i$, an Oriented Bounding Box (OBB) $b_i$, and a node category $c_i^{\text{node}} \in \mathcal{C}^{\text{node}}$. Conversely, each edge $e_{i \rightarrow j} \in \mathcal{E}$, connecting node $v_i$ to $v_j$ where $i \neq j$, is characterized by an edge category or semantic relationship denoted by $c_{i \rightarrow j}^{\text{edge}} \in \mathcal{C}^{\text{edge}}$, or can be written in a relation triplet <\textit{subject, predicate, object}>. Here, $L$, $\mathcal{C}^{\text{node}}$, and $\mathcal{C}^{\text{edge}}$ represent the sets of all entity labels, node categories, and edge categories, respectively. Given the 3D scene data $D_i$ and $D_j$ that represent the same point cloud of a scene but from different views (rotation and transition), we try to predict the probability distribution of the equivariant scene graph prediction in which the equivariance is preserved:
\begin{equation}
    \begin{cases}
        P(\mathcal{G} | D_i) = P(\mathcal{G} | D_j)_{ i \neq  j} \\
        D_j = R_{i \rightarrow j}D_i + T_{i \rightarrow j} 
    \end{cases}
\end{equation}
where $R_{i \rightarrow j}$ is the rotation matrix and $T_{i \rightarrow j}$ is the transition matrix.

\subsection{Feature Extraction} \label{sec:Feature Extraction}
In this phase, the framework extracts the feature for scene graph generation, following the three main steps: point cloud reconstruction, geometry segmentation, and point cloud extraction with nodes.

\paragraph{Point Cloud Reconstruction} The proposed framework will take the point cloud data, which can be reconstructed from various techniques such as ORB-SLAM3 or HybVIO \cite{ORBSLAM3_TRO, HybVIO}, as the input. However, for the objective validation purpose of scene graph generation, we use the indoor point cloud dataset 3RScan \cite{3RScan} for ground truth data $D_i$.
\paragraph{Geometric Segmentation and Point Cloud Extraction with Segments Nodes} Given a point cloud $D_i$, this geometric segmentation will provide a segment set $\mathbf{S}=\left\{\mathbf{s}_1, \ldots, \mathbf{s}_n\right\}$. Each segment $s_i$ consists of a set of 3D points $\mathbf{P}_i$ where each point is defined as a 3D coordinate $p_i \in \mathbb{R}^3$ and a color. Then, the point cloud concerning each entity is fed to the point encoders for node and edge features.

\subsection{Scene Graph Generation} \label{sec:Scene graph prediction}
In this phase, the framework processes the input from feature extraction (Section \ref{sec:Feature Extraction}) to generate the scene graph.

\paragraph{Properties and Neighbor Graph Extraction} From the point cloud, we extract features including the centroid $\overline{\mathbf{p}}_i \in \mathbb{R}^3$, standard deviation $\boldsymbol{\sigma}_i \in \mathbb{R}^3$, bounding box size $\mathbf{b}_i \in \mathbb{R}^3$, maximum length $l_i \in \mathbb{R}$, and volume $\nu_i \in \mathbb{R}$. We create edges between nodes only if their bounding boxes are within 0.5 meters of each other, following \cite{wu2021scenegraphfusion}.

\paragraph{Point Encoders} PointNet \cite{8099499} $f_p(\mathbf{P_i})$ encodes the segments $s_i$ into latent node and edge features, which are then passed to the model detailed in Section \ref{sec:method}.

\paragraph{Node and Edge Classification} Node classes and edge predicates are predicted using two Multi-Layer Perceptron (MLP) classifiers. Our network is trained end-to-end with a joint cross-entropy loss for both objects $\mathcal{L}_{\text{obj}}$ and predicates $\mathcal{L}_{\text{pred}}$, as described in \cite{Wald2020}.

\section{Equivariant Scene Graph Generation} \label{sec:method}
For the scene graph generation, we propose the combination of Equivariant Graph Convolution Layers \cite{pmlr-v139-satorras21a} and the Graph Convolution Layers with Feature-wise Attention \cite{wu2021scenegraphfusion} for network architecture. The overall network architecture is shown in Figure \ref{fig:egnn}, and the details of each layer are presented below. 

\begin{figure*}[htbp]
    \centering
    \includegraphics[width=\textwidth]{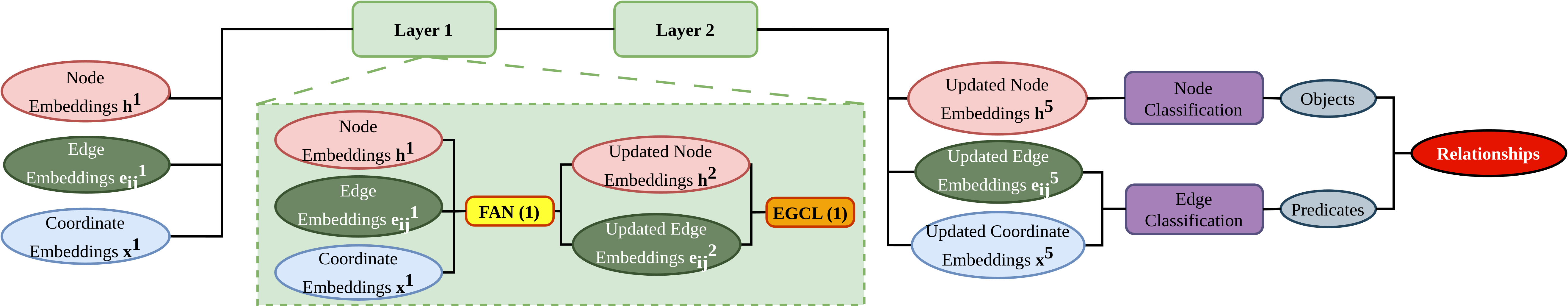}
    \caption{ESGNN Architecture.}
    \label{fig:egnn}
\end{figure*}

\subsection{Graph Initialization}
\paragraph{Node features} The node feature includes the invariant features $\mathbf{h_i}$ and vector coordinate $\mathbf{x_i} \in \mathbb{R}^3$. $\mathbf{h}_i$ consists of the latent feature of the point cloud after going through the PointNet $f_p(\mathbf{P_i})$, standard deviation $\sigma_i$, log of the bounding box size $\ln \left(\mathbf{b}_i\right)$, log of the bounding box volume $\ln(v_i)$, and log of the maximum length of bounding box $\ln(l_i)$. The coordinate vector of the bounding box $x_i$ is defined by the coordinate of the two furthest corners of the bound box. $\mathbf{h_i}$ and $x_i$ are then fed to the MLP for predicting the label of the nodes. Mathematically: 
\begin{align*}
    \mathbf{v}_i &= \left(\mathbf{h}_i, \mathbf{x}_i \right) \\
    \mathbf{h}_i &= \left[f_p\left(\mathbf{P}_i\right), \boldsymbol{\sigma}_i, \ln \left(\mathbf{b}_i\right), \ln \left(\nu_i\right), \ln \left(l_i\right)\right] \\
    \mathbf{x}_i &= [\mathbf{x}_i^{bottom right}, \mathbf{x}_i^{top left} ] \\
    c_i^{node} &= g_v\left( \mathbf{v}_{i} \right),
\end{align*}
\paragraph{Edge features} For an edge between a source node \(i\) and a target node \(j\) where \(j \neq i\), the edge visual feature $c_{i \rightarrow j}^{\text{edge}}$ is computed as follows:
\begin{align*}
\mathbf{r}_{ij} &= \left[ \overline{\mathbf{p}}_i - \overline{\mathbf{p}}_j, \boldsymbol{\sigma}_i - \boldsymbol{\sigma}_j, \mathbf{b}_i - \mathbf{b}_j, \ln\left(\frac{l_i}{l_j}\right), \ln\left(\frac{\nu_i}{\nu_j}\right) \right], \\
c_{i \rightarrow j}^{\text{edge}} &= g_e\left( \mathbf{r}_{ij} \right),
\end{align*}
where \(g_v(\cdot)\), \(g_e(\cdot)\) are MLP classifiers that project the properties into a latent space.

\vspace{-1mm}
\subsection{Equivariant Scene Graph Neural Network (ESGNN)} \label{sec:ESGNN}

Our GNN network, ESGNN, has two main components: \textcircled{1} Feature-wise attention Graph Convolution Layer (FAN-GCL); and \textcircled{2} Equivariant Graph Convolution Layer (EGCL). FAN-GCL, proposed by \cite{wu2021scenegraphfusion}, is used to handle the large input queries $Q$ of dimensions $d_q$ and targets $T$ of dimensions $d_{\tau}$ by utilizing multi-head attention. On the other hand, EGCL, proposed by \cite{pmlr-v139-satorras21a}, is used to maintain symmetry-preserving equivariance, allowing us to incorporate the bounding box coordinates \( x_i \) as node features and update them through the message-passing mechanism.

\textit{Message Passing:} ESGNN is constructed with 4 message-passing layers, consisting of 2 levels of FAN-GCL followed by the EGCL. Our model architecture is illustrated in Figure \ref{fig:egnn}, and the formula used to update node and edge features ($\mathbf{v}_i^{\ell}, \mathbf{e}_{i j}^{\ell}$) of FAN as well as the EGCL is as follows:
\begin{itemize}
    \item Update FAN-GCL:
    \begin{equation*}
        \begin{gathered}
        \mathbf{v}_i^{\ell+1}=g_v\left(\left[\mathbf{v}_i^{\ell}, \max _{j \in \mathcal{N}(i)}\left(\operatorname{FAN}\left(\mathbf{v}_i^{\ell}, \mathbf{e}_{i j}^{\ell}, \mathbf{v}_j^{\ell}\right)\right)\right]\right), \\
        \mathbf{e}_{i j}^{\ell+1}=g_e\left(\left[\mathbf{v}_i^{\ell}, \mathbf{e}_{i j}^{\ell}, \mathbf{v}_j^{\ell}\right]\right),
        \end{gathered}
    \end{equation*}
    \item Update EGCL:
        \begin{equation*}
        \begin{gathered}
        h_i^{(l+1)} = h_i^{(l)} + \text{g}_\text{v} \left( \text{concat} \left( h_i^{(l)}, \sum_{j \in \mathcal{N}(i)} e_{ij}^{(l)} \right) \right) \\
        e_{ij}^{(l+1)} = \text{g}_\text{e} \left( \text{concat} \left( h_i^{(l)}, h_j^{(l)}, \| \mathbf{x}_i^{(l)} - \mathbf{x}_j^{(l)} \|^2, e_{ij}^{(l)} \right) \right) \\
        \mathbf{x}_i^{(l+1)} = \mathbf{x}_i^{(l)} + \sum_{j \in \mathcal{N}(i)} (\mathbf{x}_i^{(l)} - \mathbf{x}_j^{(l)}) \cdot \phi_\text{coord}(e_{ij}^{(l)}) 
        \end{gathered}
        \end{equation*}
        
\end{itemize}

\subsection{ESGNN With Image Encoder} \label{sec:Image encoder}
Similar to segments $s_i$, we get the region-of-interest (ROI) from multiple corresponding images and feed it through the image encoder \cite{9578559}. Using the similar EGCL layer ensures the properties of the bounding box coordinate, we also observe better results demonstrated in Section \ref{sec:jointssg results}. The node feature is fed to node classification for object prediction and the edge feature is fed to edge classification for predicate prediction.

\captionsetup[table]{skip=6pt}
\section{Experiments} \label{sec:experiments}


 
\subsection{Dataset and Metrics}\label{sec:dataset metrics}

\paragraph{Dataset}We use the 3DSSG - a dataset for scene graph generation built upon 3RScan\cite{3RScan} which is a large-scale, real-world dataset that features 1482 3D reconstructions/snapshots of 478 naturally changing indoor environments - adapting the setting from \cite{wu2021scenegraphfusion} \footnote{https://github.com/ShunChengWu/3DSSG}. The 3RScan dataset \cite{3RScan} is processed with ScanNet \cite{scannet} for geometric segmentation. The scene graph ground truths in 3DSSG are divided into 2 versions: \textbf{l20}, which includes 20 objects and 8 predicates, and \textbf{l160}, which includes 160 objects and 26 predicates. We mainly use the test set of the \textbf{l20} version for our experiment. 

\paragraph{Metrics} Given the dataset is unbalanced \cite{Wald2020} and the objective of scene graphs is to capture the semantic meaning of the surrounding world the most, we use the \textbf{recall} of node and edge as our evaluation metrics. In the training phase, we calculate the recall as the true positive overall positive prediction. For more detailed analysis, we also adopt the metric \(\text{\textbf{R@x}}\) \cite{Wald2020, wu2021scenegraphfusion, wu2023incremental}, which takes \(\textbf{x}\) most confident predictions and marks it as correct if at least one of these predictions is correct. We apply the recall metrics for the predicate (edge classification), object (node classification), and relationship (triplet <\textit{subject, predicate, object}>). 

\vspace{-1.5mm}
\subsection{Results} \label{sec:results}

Overall, ESGNN is shown to converge more quickly in the early training epochs and achieve competitive performance throughout further epochs. Table \ref{table:main_result_l20} compares the results between our proposed model - ESGNN with existing models 3DSSG \cite{Wald2020} and SGFN \cite{wu2021scenegraphfusion} on the 3DSSG-l20 dataset with geometric segmentation setting. Ours obtains high results in both relationship, object, and predicate classification. Especially, ESGNN outperforms the existing methods in relationship prediction and obtains significantly higher \textbf{R@1} in predicate compared to SGFN. ESGNN also works well with unseen data, with competitive results compared to SGFN, shown in Table \ref{table:new_result_l20}.
\\
\begin{table}[h!]
\centering
\caption{Scene graph predictions for relationship triplet, object, and predicate, measured on 3DSSG-l20. The \textit{Recall} column reports the recall scores on objects (\textit{Obj.}) and relationships (\textit{Rel.})}
\begin{tabularx}{\columnwidth}{l c*{7}{X}}
\toprule
\multirow{2}{*}{\textbf{Method}} & \multicolumn{2}{c}{ \textbf{Relationship }} & \multicolumn{2}{c}{ \textbf{Object} } & \multicolumn{2}{c}{ \textbf{Predicate} } & \multicolumn{2}{c}{\textbf{Recall}} \\
& R@1 & R@3 & R@1 & R@3 & R@1 & R@2 & Obj. & Rel.\\
\midrule

3DSSG & 32.65 & 50.56 & 55.74 & 83.89 & $\mathbf{95.22}$ & 98.29 & 55.74 & $\mathbf{95.22}$\\
SGFN & 37.82 & 48.74 & 62.82 & $\mathbf{88.08}$ & 81.41 & 98.22  & 63.98 & 94.24\\
ESGNN & $\mathbf{43.54}$ & $\mathbf{53.64}$ & $\mathbf{63.94}$ & 86.65 & 94.62 & $\mathbf{98.30}$ & $\mathbf{65.45}$ & 94.62\\
\bottomrule
\end{tabularx}
\label{table:main_result_l20}
\end{table}
\\
\begin{table}[h!]
\centering
\caption{Scene graph predictions for new unseen relationship triplet, object, and predicate, measured on 3DSSG-l20 with geometric segmentation.}
\begin{tabularx}{\columnwidth}{l c*{5}{X}}
\toprule
\multirow{2}{*}{\textbf{Method}} & \multicolumn{2}{c}{ \textbf{New Relationship}} & \multicolumn{2}{c}{ \textbf{New Object}} & \multicolumn{2}{c}{\textbf{New Predicate} } \\
& R@1 & R@3 & R@1 & R@3 & R@1 & R@2\\
\midrule

3DSSG & 39.74 & 49.79 & 55.89 & 84.42 & $\mathbf{70.87}$ & 83.29\\
SGFN &  $\mathbf{47.01}$ & 55.30 & 64.50 & $\mathbf{88.92}$ & 68.71 & $\mathbf{83.76}$\\
ESGNN (Ours) & 46.85 & $\mathbf{56.95}$ & $\mathbf{65.47}$ & 87.52 & 66.90 & 82.88 \\
\bottomrule
\end{tabularx}
\label{table:new_result_l20}
\end{table}

Figure \ref{fig:esgnn_recall_results} reports the recalls for nodes (objects) and edges (relationships) during training between ESGNN and SGFN on both train and validation sets. The recall slope of ESGNN in the first 10 epochs (5000 steps) is significantly higher than that of SGFN. This indicates that ESGNN has faster convergence and higher initial recall.


ESGNN consistently outperforms the pioneering works overall, or is more data-efficient than SGFN, as it does not need to generalize over rotations and translations of the data, while still harnessing the flexibility of GNNs in larger datasets.

\begin{figure}[htbp]
    \captionsetup{farskip=0pt,nearskip=4pt}
    \subfloat[Edge recall on evaluation set 3DSSG-\textbf{l20}]{\includegraphics[width=0.49\columnwidth]{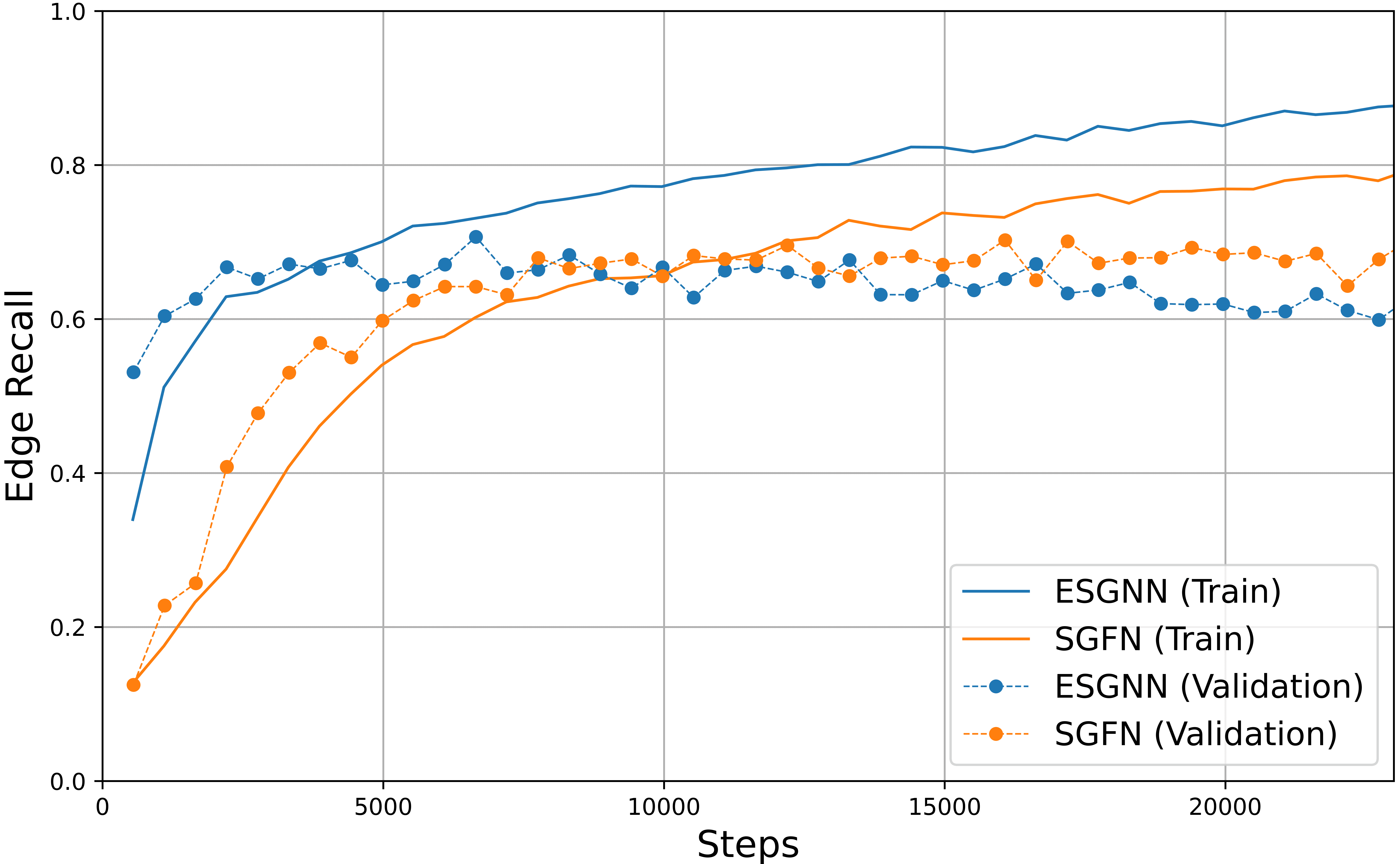}}%
    \subfloat[Node recall on evaluation set 3DSSG-\textbf{l20}]{\includegraphics[width=0.49\columnwidth]{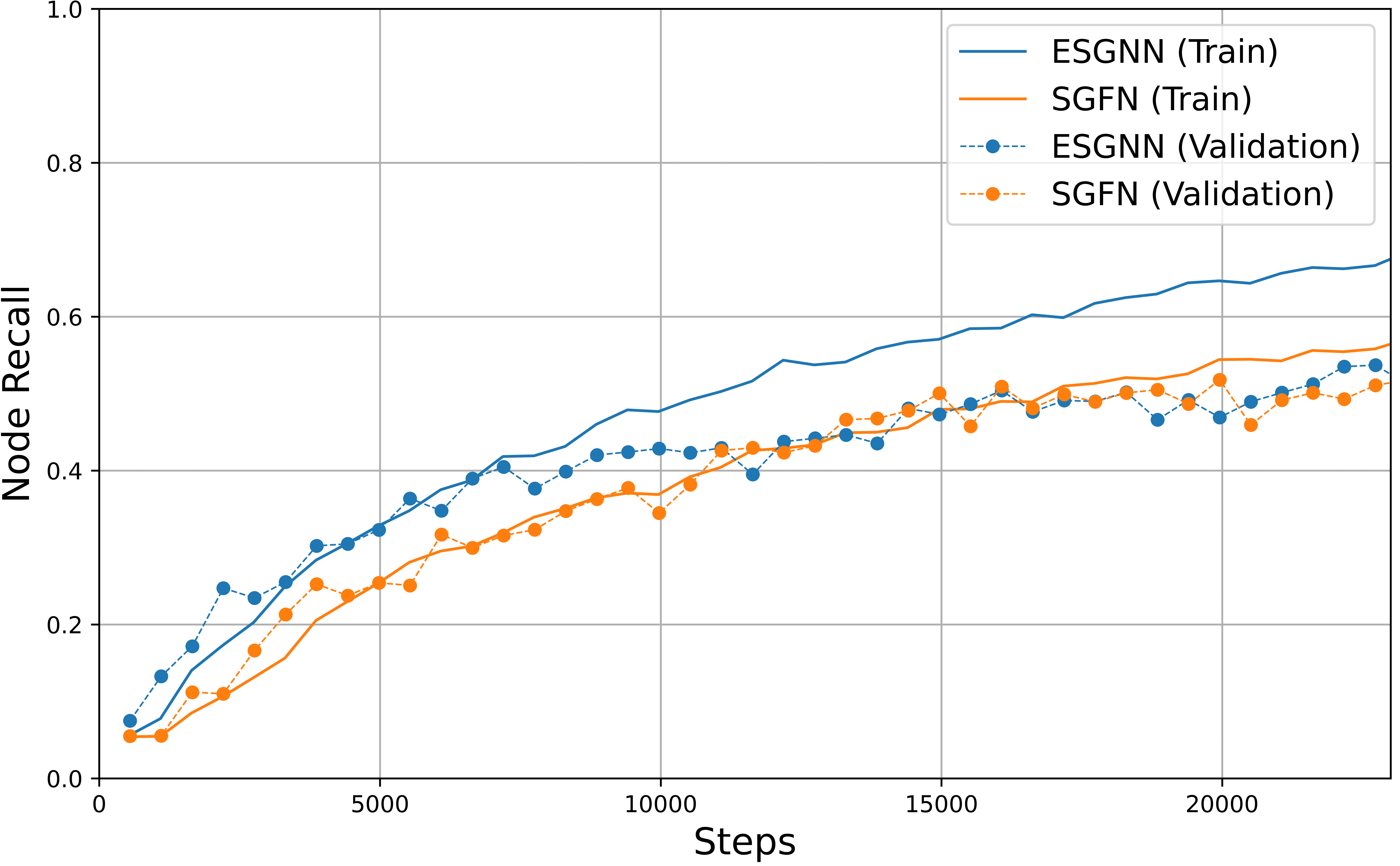}}%
    \caption{Comparison of ESGNN with SGFN through the training steps.} \label{fig:esgnn_recall_results}
\end{figure}



\subsection{Ablation Study} \label{sec:Ablation Study}

In this Section, we report the results of ESGNN with different architectures and settings that we tried, shown in Table \ref{table:main_result_l20_all}. \textcircled{1} is the SGFN, run as the baseline model for comparison. \textcircled{2} is the ESGNN with a single FAN layer and an EGCL layer. This is our best performer and is used for experiments in Section \ref{sec:results}. \textcircled{3} is ESGNN with 2 FAN layers and 2 layers EGCL. \textcircled{4} is similar to \textcircled{1}. The only difference is that we concatenate coordinate embedding to the output edge embedding after message passing. We expect this modification to improve the performance of edge prediction. \textcircled{5} is similar to \textcircled{4} with 2 layers of FAN GConv and 2 layers of EGCL.
\\
\begin{table}[h!]
\centering
\caption{Evaluation of different ESGNN architectures on scene graph generation task on 3DSSG-l20 dataset. \textcircled{2} is our best performer and is used for the evaluation in Section \ref{sec:results}}
\begin{tabularx}{\columnwidth}{l*{6}{X}}
\toprule
\multirow{2}{*}{\textbf{Method}} & \multicolumn{2}{c}{\textbf{Relationship}} & \multicolumn{2}{c}{\textbf{Object}} & \multicolumn{2}{c}{\textbf{Predicate}}  \\
 & R@1 & R@3 & R@1 & R@3 & R@1 & R@2 \\
\midrule
\textcircled{1} SGFN & 37.82 & 48.74 & 62.82 & $\mathbf{88.08}$ & 81.41 & 98.22  \\
\textcircled{2} ESGNN\_1 & $\mathbf{42.30}$ & $\mathbf{53.30}$ & $\mathbf{63.21}$ & 86.70 & 94.34 & $\mathbf{98.30}$ \\
\textcircled{3} ESGNN\_2 & 35.63 & 44.63 & 57.55 & 84.41 & 93.93 & 97.94 \\

\textcircled{4} ESGNN\_1X & 34.96 & 42.59 & 57.55 & 86.18 & 92.68 & 98.08 \\
\textcircled{5} ESGNN\_2X & 37.94 & 50.58 & 59.97 & 85.23 & $\mathbf{94.53}$ & 98.01 \\
\bottomrule
\end{tabularx}
\label{table:main_result_l20_all}
\end{table}

Figure \ref{fig:esgnn_recall_results} reports the edge and node recalls comparison during training. Models \textcircled{3} and \textcircled{5} perform well on the train set but poorly on the validation and test sets, potentially suffering overfitting as they contain more layers. Models \textcircled{4} and \textcircled{5} result in higher edge recalls in several initial epochs, but experience a decline in recall in the later epochs. 

\begin{figure}[htbp]
    \captionsetup{farskip=0pt,nearskip=4pt}
    \subfloat[Edge recall on evaluation set 3DSSG-\textbf{l20}]{\includegraphics[width=0.48\columnwidth]{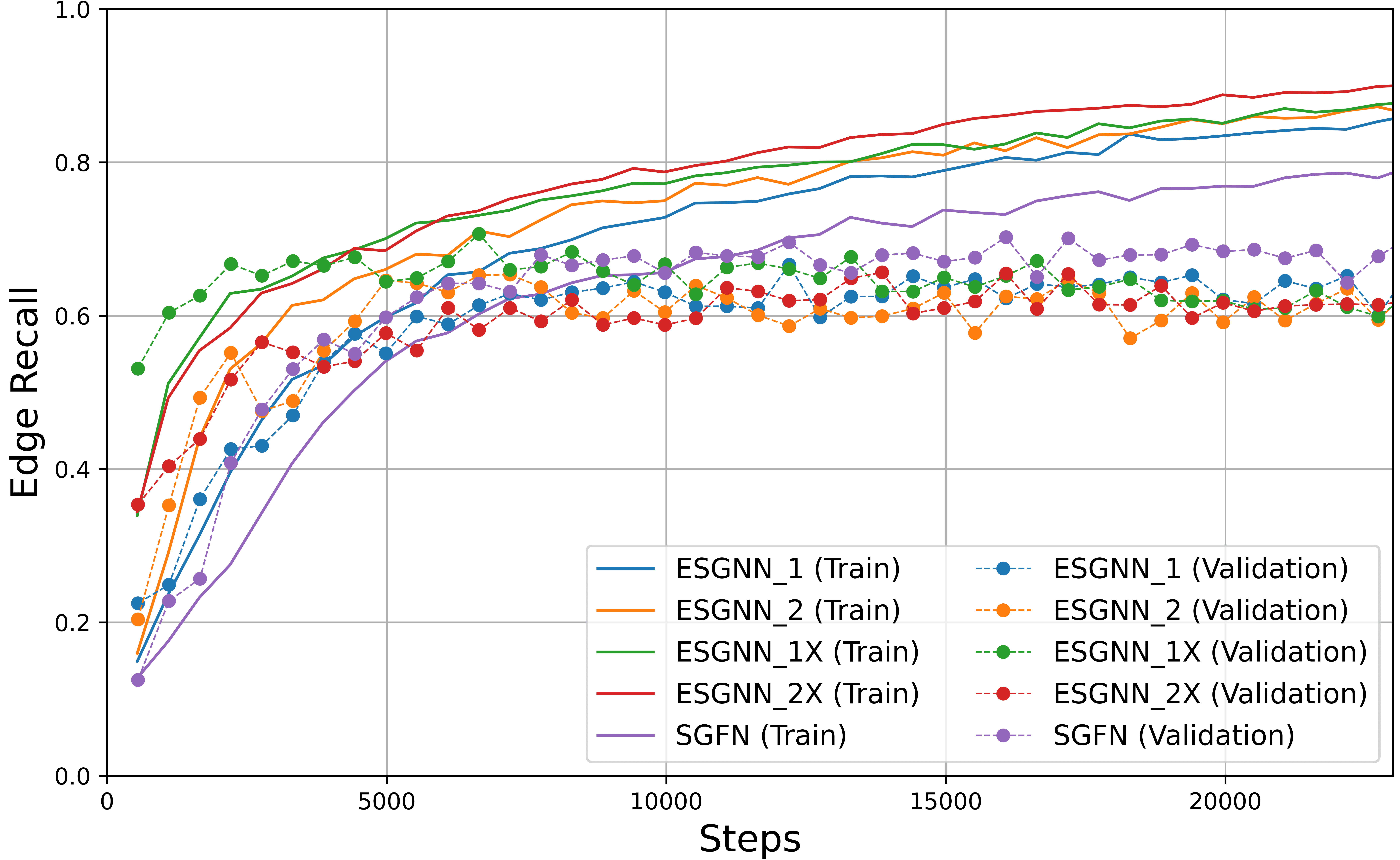}}%
    \subfloat[Node recall on evaluation set 3DSSG-\textbf{l20}]{\includegraphics[width=0.48\columnwidth]{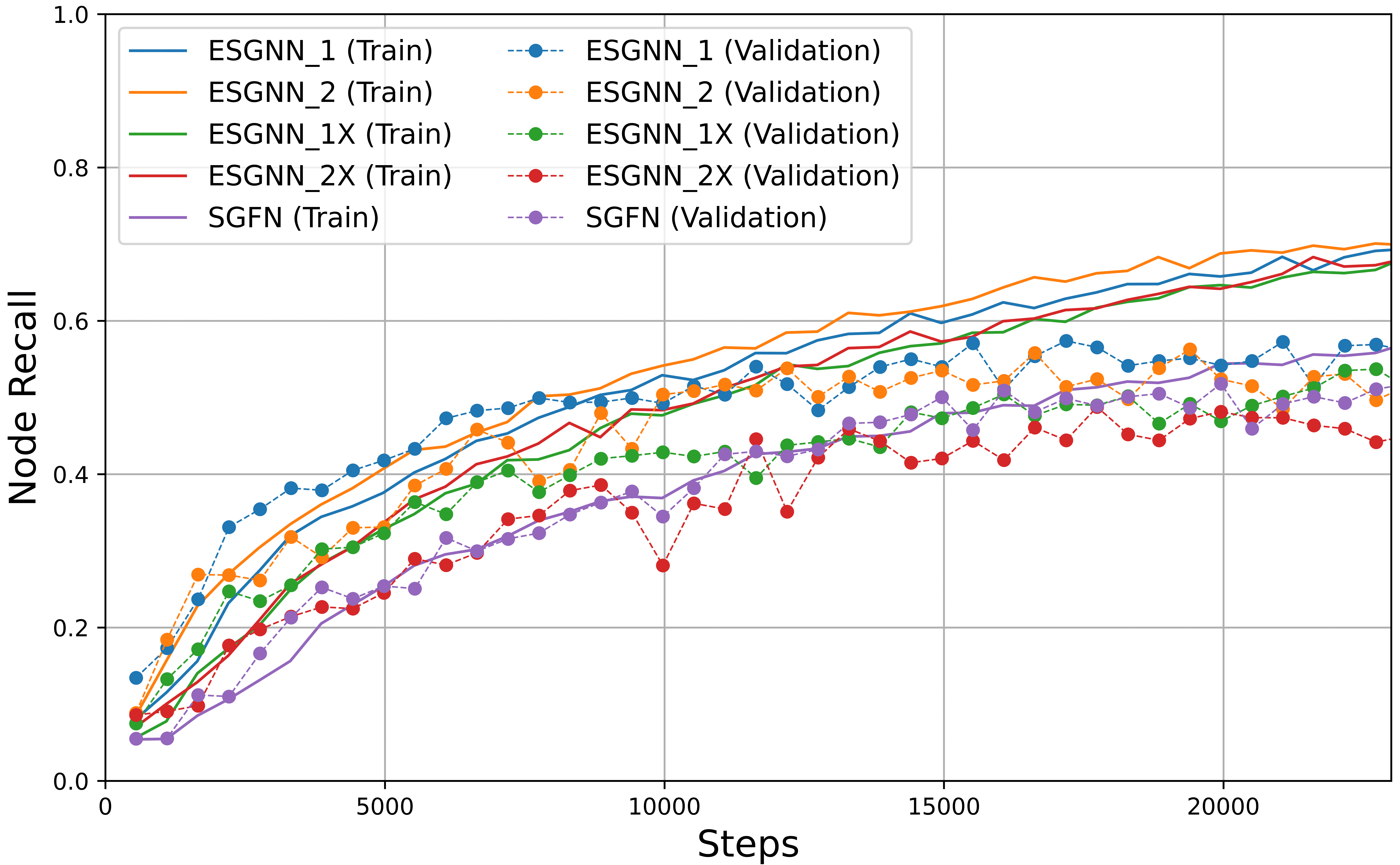}}%
    \caption{Comparison of multiple ESGNN models with SGFN through the training steps.} \protect\label{fig:esgnn_recall_results}
\end{figure}

\subsection{ESGNN with Image Encoder} \label{sec:jointssg results}
\begin{figure}[htbp]
    \captionsetup{farskip=0pt,nearskip=4pt}
    \subfloat[Edge recall on evaluation set 3DSSG-\textbf{l20}]
    {\includegraphics[width=0.48\columnwidth]{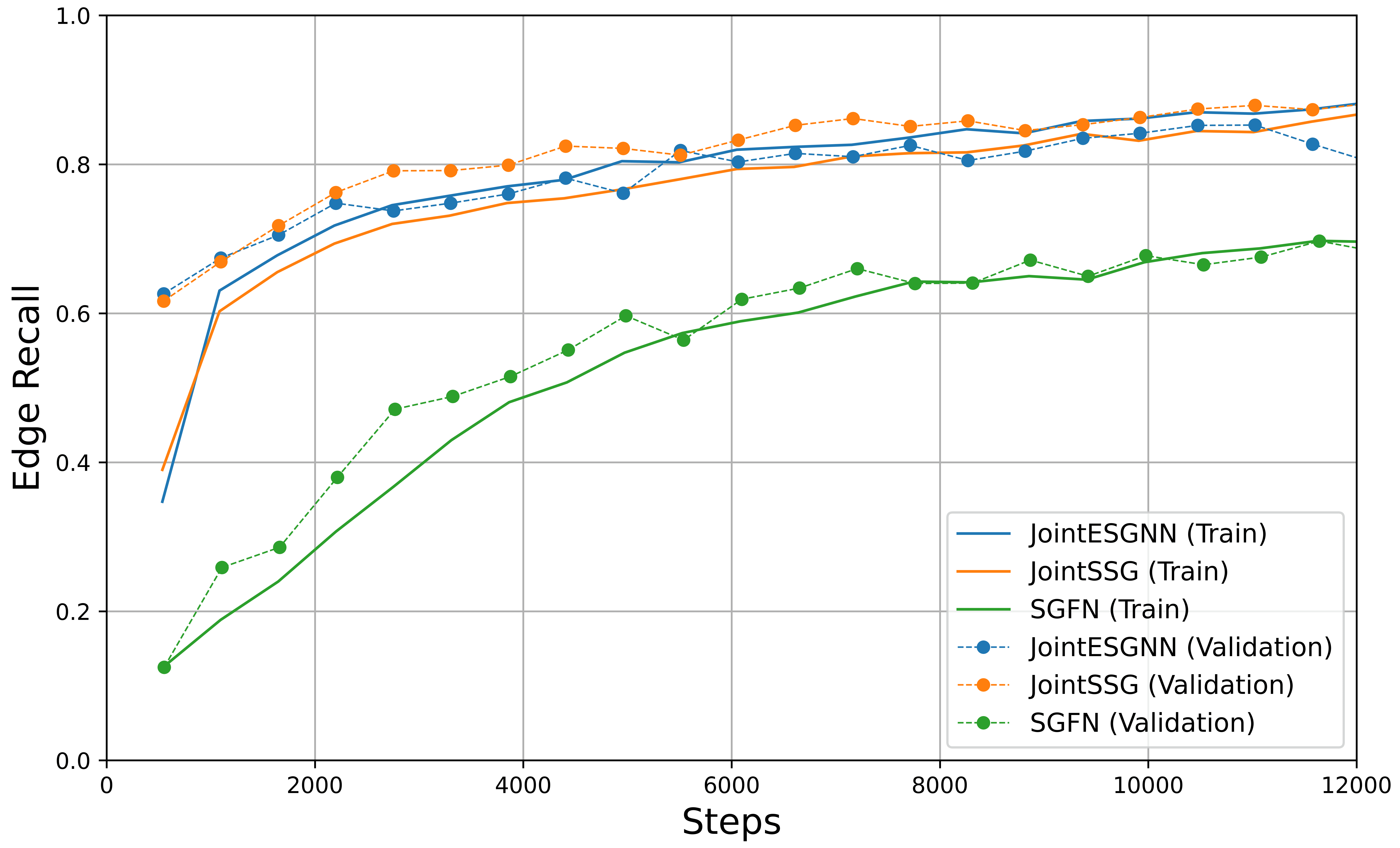}} %
    \subfloat[Node recall on evaluation set 3DSSG-\textbf{l20}]
    {\includegraphics[width=0.48\columnwidth]{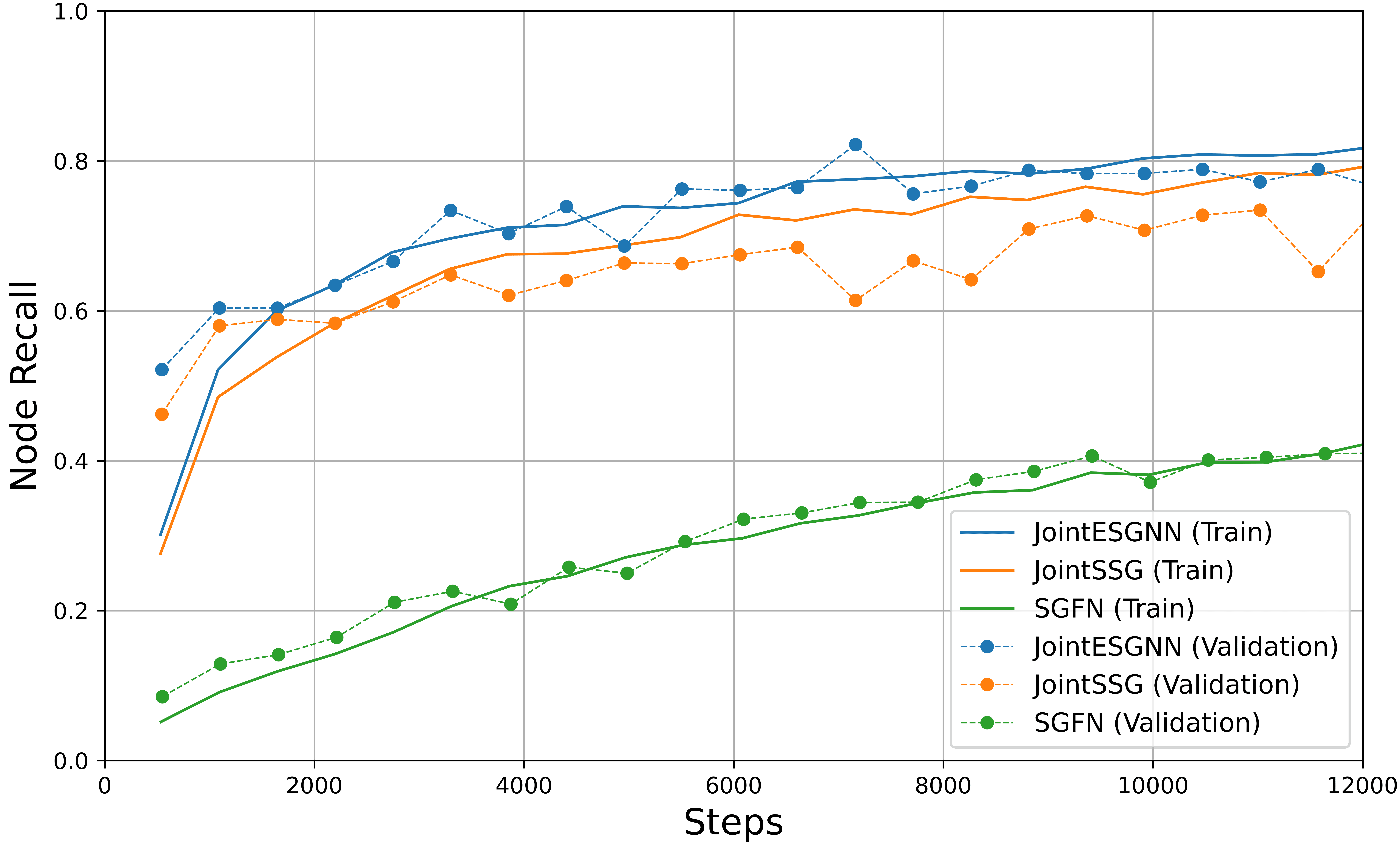}} %
    \caption{Comparison of Joint-ESGNN, SGFN, JointSSG through the training steps.} \label{fig:esgnn_recall_results_joint}
\end{figure}

Our model also poses a potential in application on point-image encoders model together such as JointSSG \cite{wu2023incremental}. We implement our GNN architecture similar to JointSSG and name it Joint-ESGNN. Figure \ref{fig:esgnn_recall_results_joint} shows the performance of our model with image encoders compared to JointSSG and SGFN. 

\section{Conclusion} \label{sec:conclusion}
In this work, we introduced the Equivariant Scene Graph Neural Network (ESGNN), which enhances robustness and accuracy in generating semantic scene graphs from 3D point clouds. Leveraging E(n) \textbf{E}quivariant \textbf{G}raph \textbf{N}eural \textbf{N}etwork (EGNN), ESGNN maintains symmetry-preserving properties, outperforming state-of-the-art methods with fewer layers and reduced computational resources. Our results demonstrate ESGNN's superior performance in generating consistent and reliable scene graphs, paving the way for more efficient 3D scene understanding frameworks in autonomous systems. Future work will optimize ESGNN for specific use cases, incorporate additional sensor data, and handle more complex scenarios.

\bibliographystyle{unsrt}
\bibliography{paper}

\vspace{12pt}

\end{document}